# 3-Survivor: A Rough Terrain Negotiable Teleoperated Mobile Rescue Robot with Passive Control Mechanism

Rafia Alif Bindu[1,2], Asif Ahmed Neloy[1,2], Sazid Alam[1,2], Shahnewaz Siddique[1]

*Abstract*— This paper presents the design and integration of 3-Survivor: a rough terrain negotiable teleoperated mobile rescue and service robot. 3-Survivor is an improved version of two previously studied surveillance robots named Sigma-3 and Alpha-N. In 3-Survivor, a modified double-tracked with caterpillar mechanism is incorporated in the body design. A passive adjustment established in the body balance enables the front and rear body to operate in excellent synchronization. Instead of using an actuator, a reconfigurable dynamic method is constructed with a 6 DOF arm. This dynamic method is configured with the planer, spatial mechanism, rotation matrix, motion control of rotation using inverse kinematics and controlling power consumption of the manipulator using angular momentum. The robot is remotely controlled using a handheld Radio Frequency (RF) transmitter. 3-Survivor is equipped with a Raspberry Pi 12 MP camera which is used for live streaming of robot operations. Object detection algorithms are run on the live video stream. The object detection method is built using a Faster R-CNN with VGGNet-16 architecture of CNN. The entire operations of the robot are monitored through a web control window. Therefore, the control portal provides a brief scenario of the environment to run, control and steer the robot for more precise operation. A very impressive 88.25% accuracy is acquired from this module in a rescue operation. Along with the ODM, the sensor system of the robot provides information on the hazardous terrain. The feasibility of the 3-Survivor is tested and presented by different experiments throughout the paper.

## I. INTRODUCTION

### A. Motivation

Currently, robots are currently working as human assistants or as an alternative to human operators in diverse activities. At this time, the uses of robots are greatly increasing in different military operations like rescuing, surveillance, defeating terrorist indignations, detecting enemies, bomb defusing [1-4]. Rescue robots were first introduced in 1986 at the Chernobyl nuclear accident in Ukraine [5]. Currently, rescue robots are developing at an unrivaled pace. In fact, many rescue robots are researched and developed to perform in both hazardous and indoor environments. Rescue robot in such environment includes tracked vehicles [6], aerial robots [7], snake-like robots [8] and so forth. For using the robots in hazardous areas, the robot must have high mobility ability to move in rough terrain, clearing the path and also transmitting live video of the affected area. Generally, three types of robot locomotion are found. The first is the walking robot [9], second, the wheel-based robot [10] and the third is a track-based robot [6]. The wheel-based robot is not efficient for moving in the rough terrain [11]. The walking robot can move but it's very slow and difficult to control [9]. In that sense, the double-track mechanism provides good mobility under rough terrain conditions with the added benefit of low power consumption. Considering all these conditions, the track-based robot is more efficient in rough terrain as well as in a rescue operation [12]. The main aim of the study is to make the rescue operation more efficient. The rescue robot operates alongside the traditional rescue team and makes the rescue operations faster and more effective in various hazardous areas.

### B. Construction of 3-Survivor

Compared with developed countries, the research on rescue robots in Bangladesh started very late. Some of the rescue or surveillance robots are developed to perform light-hand operations [13]. Most of them are not capable of handling hazardous environments or surveillance activities [13]. In this study, the authors are proposing 3-Survivor, a new version of Sigma-3 [14] and Alpha-N [15], for performing potential operations in rescuing or surveillance under both indoor and hazardous environment. In particular, 3-Survivor is the improved and combined version of the previous two. The first version of the robot, Sigma-3, was a service robot that was developed to execute operations like holding, grabbing and transferring objects by using a 6 DOF Robotics arm. The operation is functioned by live streaming operated by a Raspberry Pi 0.6 MP camera. On the other hand, the second version of Sigma-3, the Alpha-N robot, utilizes live streaming for object detection and autonomous movement with Shortest Path Finder using Grid Count Algorithm, object or obstacle detecting module using Faster R-CNN with VGGNet-16. The obstacle avoiding algorithm is built with an improved Dynamic Window Approach (IDWA) and Artificial Potential Field (APF). So, 3-Survivor is the final integrated version of both Sigma-3 and Alpha-N that exhibits the full functionality of a rescue robot. In particular, the overall feasibility of the robot is tested at an accredited test

[1] Electrical and Computer Science Department, North South University, Plot, 15, Block B Kuril - NSU Rd, Dhaka 1229, Bangladesh

[2] TF-ROS Lab, North South University

Emails: {rafia.bindu, asif.neloy, sazid.alam, shahnewaz.siddique}
@northsouth.edu

environment. Besides, additional experiments were conducted to enhance the on-test applicability.

C. *Organization of this paper*

The paper is organized as follows: In section III.A, the design of the double-track mechanism is presented. Section III.B presents the integration methods of Object Detection Module (ODM). In section III.C the sensor system on 3-Survivor is presented and previously adopted methods [14] are studied and explained. Finally, the performance of the 3-Survivor is shown in section IV by simulation and experiments at different environments and velocities on testing ground.

I. LITERATURE REVIEW

A. *Problem Statement*

Rescue Robots are very crucial for immediate response and assessment of the environment in a situation like natural disasters or man-made complex surroundings. But it is difficult to perform missions autonomously in these emergencies by rescue robots. In particular, the role of rescuing persons or surveilling the situation during combat is also often performed by trained soldiers in action. A similar situation occurred in the Holey Artisan Bakery Café in Dhaka [16]. A group of expert commandos rescued the survivors. There were no robots present to assist in the situation and ease the operations. On the contrary, these kinds of situations can be easily handled by rescue and surveillance robots [4]. YTA ScoutBot [17], India's Riot-Control Drones [18], South Korea's Prison Robo-Guards [18], Israel's Deadly Rover [18], Brazil's Olympic Peacekeepers [18] are examples of such robots which are currently in use and perform state-of-art research. The primary objective of this research is to develop a rescue as well as a service robot for multi-purpose operations that may be especially carried out in Bangladesh. Along with rescue jobs, the robot should be able to perform the duty of basic service robots like helping humans to transfer an object from one place to another, survive any environment, detecting an object or any particular human to work as a spy bot, etc. These types of services will increase the usability of these robots in developing countries like Bangladesh.

B. *Proposed Solution*

The design and test method of a rescue robot largely depends on the reliability and risk assessment of the robot. For active and reliable rescue operations, the design of 3-Survivor is split into 3 parts - mechanical part, software part, and humanoid control part. Maximum attainable risks [19] are proposed to minimized in 3-Survivor from existing state-of-arts rescue robots through the following proposals -

1. Systematic mechanical design faults are minimized by the proposed double-tracked mechanism having Fault Tree Analysis (FTA) [19].
2. For software and random component failures, previous versions of 3-Survivor are deployed in testing phases with performance measurement.
3. Human and miscellaneous errors are tested in 3 phases for minimizing errors - independent move, terrain move, speed control. All the simulations are verified and debugged several times until maximum performance is observed.

In rescue operations, time management and observing the surrounding environment is an important task for managing efficient operations. Most rescue robots are aware of ground conditions using temperature, humidity, ultrasonic sensor, and 360-degree movable flipping cameras. In conjunction with these existing sensors, 3-Survivor incorporates additional sensors for supporting operations including HC-SR04 Ultrasonic Sonar Sensor, DHT22 Temperature and Humidity Sensor, HMC5883L 3-Axis Compass, MQ 02 Gas Sensor and Raspberry Pi 12MP Camera integrated with an Object Detection Module (ODM). The sensor system and object detection module increase the sustainability of the diagnostic and navigation system for the robot in robust and unknown environments such as chemical factories, bomb defusal, surveillance of hazardous environments where 3-Survivor can replace humans in these high-risk operations. Also, the live footage of the pi camera is displayed to a web interface to facilitate the operations closely as a part of teleoperation. Moreover, the sensor system is constructed with a minimum sensor array to minimize computational complexity. Since dynamic forces affect the motion of the base and the manipulators in a mobile robotic system, the sensor system along with the camera is distributed along the body to maintain passive adaptivity while moving in rough terrain. Thus, the body design mechanisms in 3-Survivors have been substantially updated from the previous version Sigma-3. Finally, object detection is very important for any rescue robot to detect objects where human eyes couldn't reach. Most rescue robots require extensive object detection software to identify humans in danger. Some of the major studies already presented include models implemented using FPGA [20], 3D LIDAR [21], kinetic-based vision system [22] and sensor-based vision [23]. Many rescue robots also use the mentioned features for automatic navigation and path planning [24]. However, 3-Survivor primarily addresses the problem in the context of a robot exploring an unknown environment with the goal of finding causalities or objects and a supplementary method for the human to understand the environment for making further decisions. Therefore, the accuracy and performance of the ODM, which was studied and implemented in Alpha-N, is improved significantly in 3-Survivor.

The key contribution of this paper is to collect the best practices and combine the different architectural approaches in a rescue robot. The main contributions of this paper are summarized below:
1. 3-Survivor is specially built for developing countries like Bangladesh, where there is a large demand for rescue robots but at present, it is rare to use a robot for the rescue operation.
2. Performance of Sigma-3 and Alpha-N is significantly improved in 3-Survivor with both improvements in design and architecture.
3. The double-tracked caterpillar mechanism is improved from existing studies [12]. This is improved

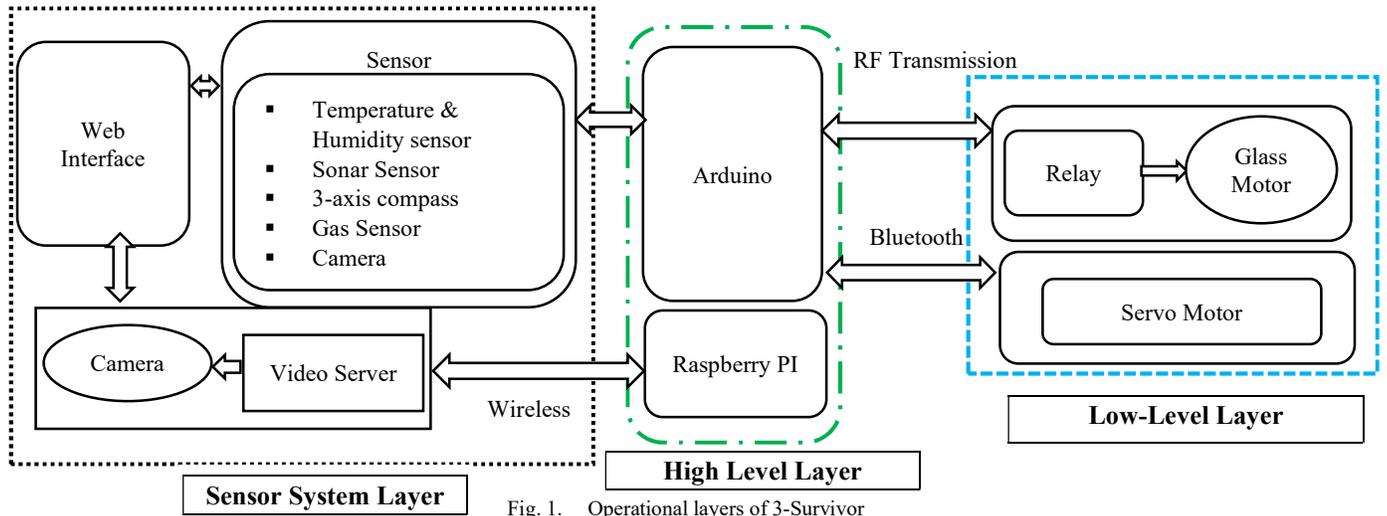

Fig. 1. Operational layers of 3-Survivor

4. by smoothing the vibration absorption system for better body mass distribution.
5. Object detection methods are reconstructed and redeployed especially for support control and rescue system which is absent in most of the rescue and surveillance robots.

## II. THE ARCHITECTURE OF 3-SURVIVOR

As shown in Figure 1, the architecture of 3-Survivor consists of three layers-

1. A high-level layer transmits and receives data from the sensor system. This layer consists of Raspberry Pi 3B+ and Arduino Uno R3. The controller encodes and sends data to the low-level layer to control each motor.
2. A low-level layer that runs all the actuators and motors in 3-Survivor and maintains communication with the high-level.
3. The sensor system layer.

On the other hand, system integration has 2 main sites- robot site and station site. The station site is the robot operator's a zone where the robot is controlled by a 6CH FS-CT6B radio. The controller and the web interface are connected by a Raspberry Pi WIFI module. To lower the complexity of the processing unit, Raspberry Pi and Arduino are connected via the I2C bus as master and slave. It also enables data processing with a high-level layer to the lower-level layer.

All the configuration and body details are stated in Table 1.

Table 1. Configuration of 3-Survivor

| Specification | Configuration/Measurement | |
|---|---|---|
| Material | Lightweight aluminum | |
| Weight Distribution (Without Object) | Item | Weight (Kg) |
| | 6 DOF Arm (including 7 Servo) | 4.3 |
| | Front & Rear Track | 18.70 |
| | Others | 0.63 |
| | | Total = 23.63 ~ 24 |
| Motors | 12V Glass Motor | |
| Wheels | Caterpillar track with a single chain | |
| Sensors | Item | Model |
| | Temperature & Humidity | DHT22 |
| | Ultrasonic Sonar | HC-SR04 |
| | Gas Sensor | MQ-2 |
| | 3-axis compass | HMC5883L |
| Camera | Raspberry Pi Camera Module V2 | |
| Controller | FS-CT6B | |
| Shape | Size | Unit (mm) |
| | Length | 450 |
| | Width | 270 |
| | Height | 210 |

### A. Improved Passive Double Track Mechanism

According to the numbers of the tracks adjoined on a robot, the rescue robots can be classified as one to four tracked robots. Single-tracked robots are equipped with fixed weight center or shaft in the body frame. While moving on the rough terrain, the fixed shaft affects the stability and body balance. Therefore, the single-tracked mechanism is not suitable for rescue or surveillance robot. Other tracked robots have also a limitation in high or rugged terrain due to imbalance with the wheel and passive adaptivity [12]. Considering the challenging scenario, the authors propose a novel double-tracked mechanism with passive adaptivity. This proposed methodology significantly improves the ability to navigate in the rough terrain with passive mobility which is one of the main contributions of 3-Survivor. For adapting this mechanism, the body of 3-Survivor is divided into 2 parts. The front part contains the 45° declined adaptive slope which is controlled by two glass motors for lifting (Fig.3 and Fig.4). So, the double-tracked chain can climb up to 40°slope or stair and the maximum designed payload capacity is 12KG. The rear part is incorporated with four wheels for movement.

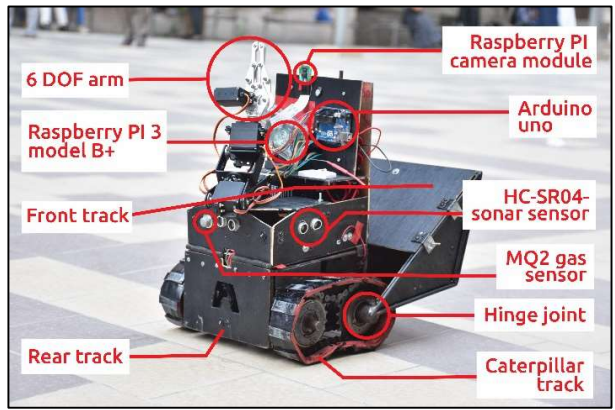

Fig. 2. The overall design of 3-Survivor

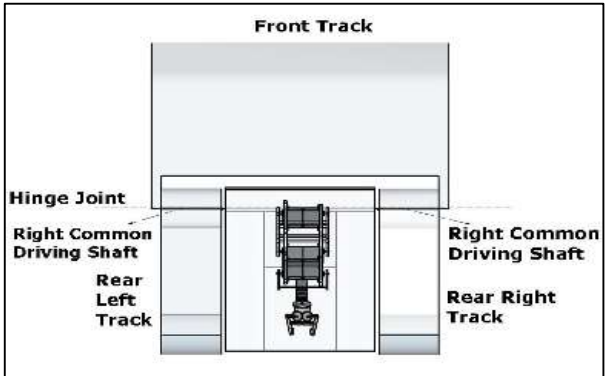
Fig. 3. Top View

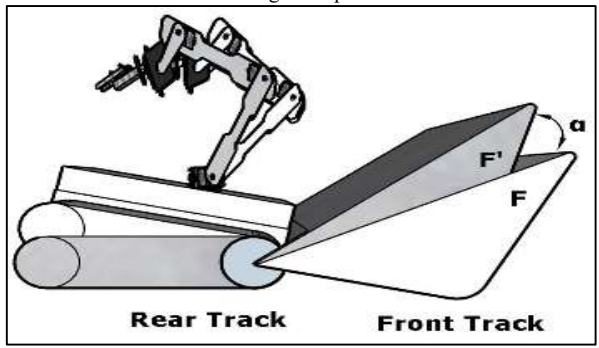
Fig. 4. Side View

The front and the rear part are connected with a fixed shaft (marked in Fig.3, Fig.4). This shaft is also a common axis for the joint of front motors with the same hinge axis (marked in Fig.3, Fig.4). Thus, the front body simultaneously allows movement (From state $F$ to $F'$) with the same rotation ($\alpha$) and direction with the rear body in both driving and reverse direction (Fig.7, Fig.8). Previously studied double-tracked mechanism only allowed two tracks at each side to rotate in the same direction as that of the driving shaft [25]. Thus, changing the weight balance configuration substantially stabilized the proposed mechanism. The adaptivity of the front and rear body established balance by moving the center of mass and Zero Moment Position (ZMP)(Fig.8).

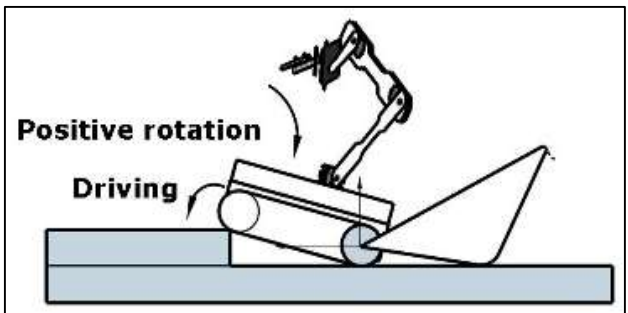
Fig. 5. Positive Rotation

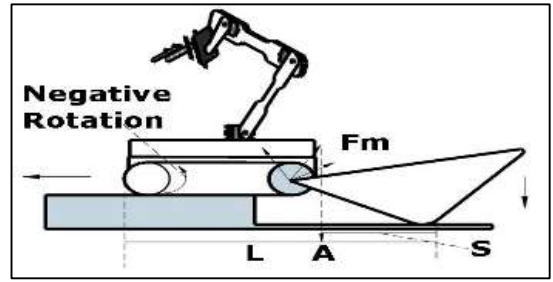
Fig. 6. Negative Rotation

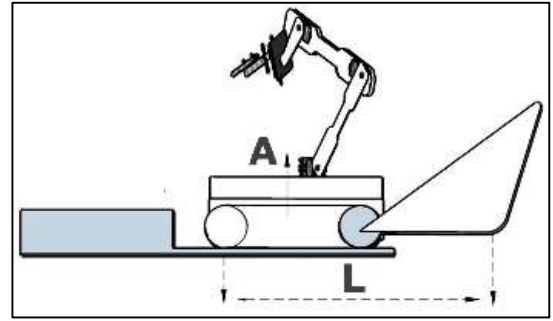
Fig. 7. Mass Deviation (Initial State)

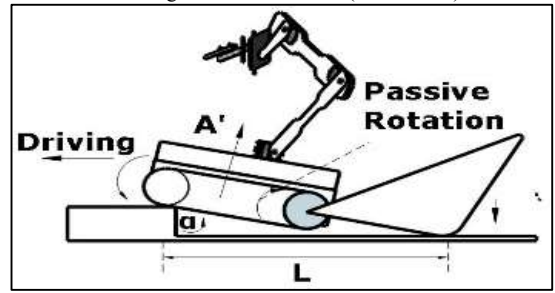
Fig. 8. Mass Deviation (Moving State)

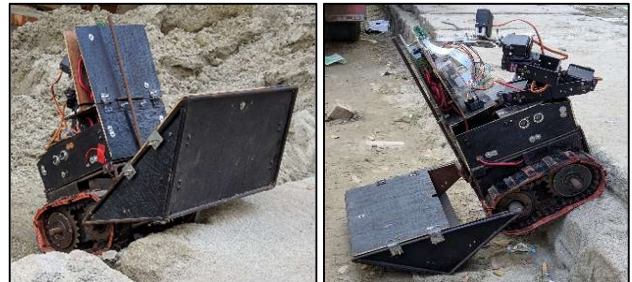
Fig. 9. Climbing test on high ground

The feasibility of this method is investigated in an example of terrain climbing as shown in Fig. 9. For better understanding, the whole process is illustrated in a 2D drawing model with positive and negative deviation (Fig. 5, Fig.6). While roaming the plane ground, the center of mass initially stays stable at $A$ and the maximum stability margin points towards $\delta$ (Fig.7, Fig.8). In the rugged grounds, the center of mass moves towards delta and changes the initial position to $A'$. Due to the track adaptability, the supporting area increases and the maximum stability margin stays the same as the initial state. The whole process is also precisely verified by stair climbing experiments (see Fig. 11).

## B. Object Detection Module (ODM)

ODM along with the sensor system is a part of the surveillance module in 3-Survivor. The ODM proposed here is a substantial improvement from the previous version, Alpha-N. A brief architectural structure of the algorithms with the experimental setup is adopted from Alpha-N. In the previous study, ODM of Alpha-N was unable to operate in the following situations -

1. Failed to detect blurry objects in uneven terrain with egomotion in some scenarios.
2. Required more computational memory to allocate the ODM, therefore the system interruption occurred frequently while performing in higher FPS.
3. Testing and validation accuracy varied from on-field streaming accuracy.

In this study, some of the inner modular setups are modified with the validation process to minimize the above lackings. To do so, the following measures are taken -

1. An extra convolutional layer is added in the Region Proposal Network (RPN) to support the Classes and Bounding Boxes Prediction.
2. The automatic pathfinding system is removed from ODM to lower the complexity. 3-Survivor is a teleoperated robot, thus automatic pathfinding is not required.
3. The training and validation dataset is improved by adding custom object classes with a custom video dataset having features described in Table 2. Also, the validation method is changed according to the environments.

Table 2. Custom Dataset for ODM

| Dataset | Nature | Testing Feature |
|---|---|---|
| D1 | Fast-moving blurred objects | Fast movement detection |
| D2 | Fast and Slow rotation of an object | Expend the frame rate for the model |
| D3 | Same object with small and large size | Correct the error rate for mismatch detection |
| D4 | Lower pixel image | Improve the partial translation rate |

Limited hardware was used to train the model with the previous procedure followed described in Alpha-N with evaluation methods. As the robot is designed to do rescue operations, test arenas are needed to provide a realistic and challenging environment for measuring overall performance. A similar setup is explained in the study of A. Jacoff, et al [26]. Details result and performance analysis are described in Section. IV

## C. Control System Integration

The sensor and control system is integrated with both 6 DOF arm and RF control system. Fig.10 shows the control web version for the operator. To enable the environmental assessment system, the web portal provides detailed information from sensors. All the sensors are connected with the lower architecture of the 3-Survivor. Thus, the processing power doesn't affect the control or operating mechanisms.

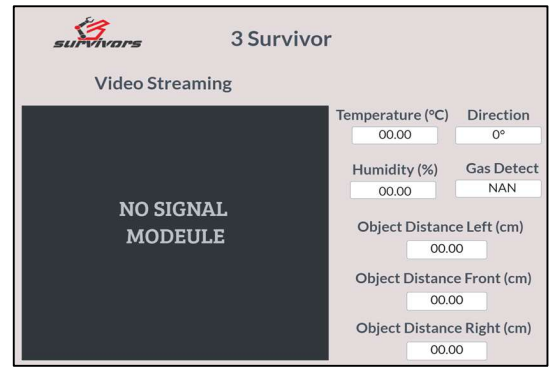

Fig. 10. Central web control of 3-Survivor

## III. DEMONSTRATION AND FEASIBILITY TEST

Results and analysis of practical demonstration by field experiments are presented in this section to investigate the actual feasibility of the proposed extraction strategy. A sample rescue operation was designed to test the prototype. The operation was designed as – A human/object is trapped in a room where the robot has to climb a stair, observe the environment with the pi camera and sensors, finally execute the rescue mission by teleoperated control of the rescue team. For a better assessment, the test took several times until the maximum performance was acquired.

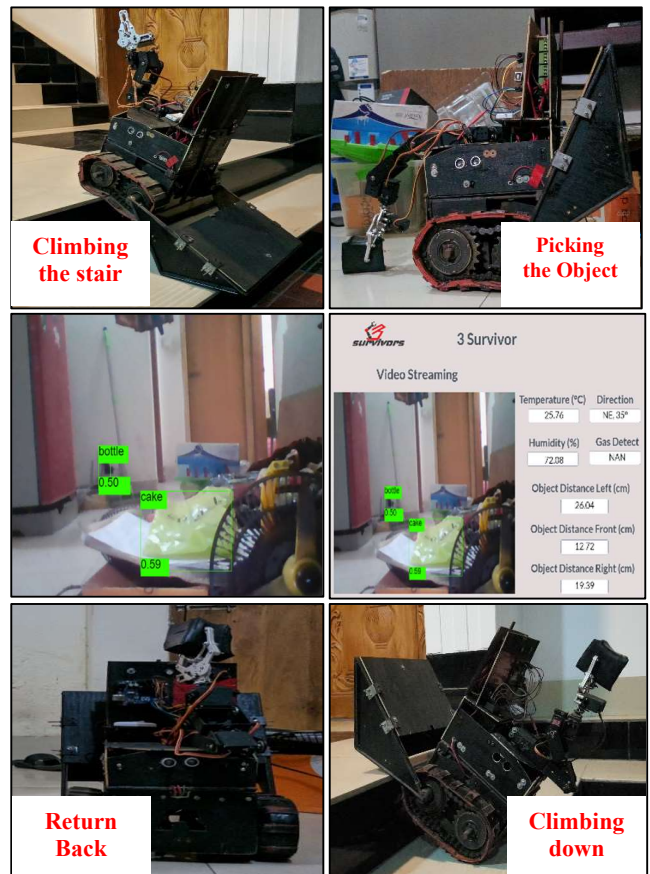

Fig. 11. Execution of a dummy operation.

Through these tests from Fig.11, a number of functionalities were evaluated – feasibility of the body design in a hazardous environment, examining operational accuracy and performance of ODM and other control systems through web portal debugging. For better understanding, the operational entities and their results are illustrated in Fig.11 to Fig.14.

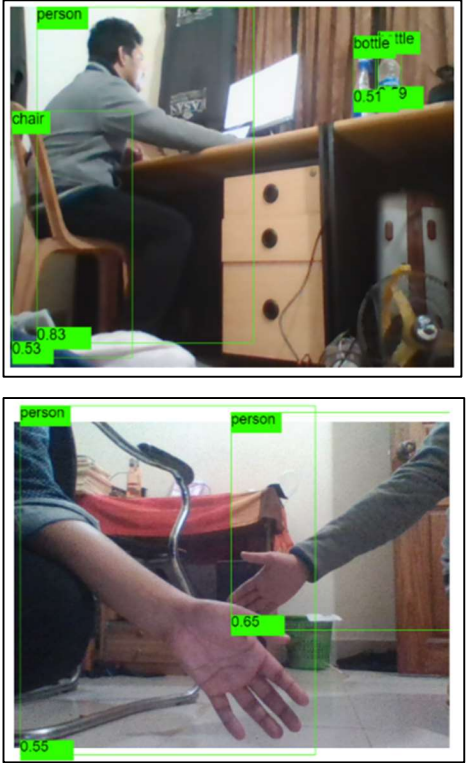

Fig. 12. Testing the object detection module

Finally, for performance evaluation of ODM, a method published in [26] was used. The results were then evaluated using the metrics of [26]. The acquired results, accuracy and model loss status (Fig.13) are described below. From the observation, the optimum score obtained from the ODM is in dataset D1 and D4 (Table 3 and Fig.13, Fig.14). The overall accuracy of the ODM is 88.25% (Dataset D) with MAP which is substantially improved from the previous version as well as compared to other state-of-art rescue robots [4, 11, 12]. Moreover, from the results, it proves that the ODM can be further improved by using fast- and slow-moving frame rate minimization techniques. Fig.13 shows the accuracy and Fig.14 shows the loss of the ODM with comparison from a different dataset.

Table 3. ODM result and validation

| Method | Formulation | Result | | | |
|---|---|---|---|---|---|
| | | Dataset | Score | AVG FPS | Accuracy |
| Recall | $\frac{TP}{TP+FN}$ | D1 | 55.6 | 17 | 83.26 |
| | | D2 | 48.9 | 13 | 81.25 |
| | | D3 | 47.1 | 12 | 80.95 |
| | | D4 | 54.9 | 16 | 82.55 |
| Precision | $\frac{TP}{TP+FP}$ | Dataset | Score | AVG FPS | Accuracy |
| | | D1 | 58.9 | 15 | 85.20 |
| | | D2 | 42.3 | 12 | 79.10 |
| | | D3 | 41.0 | 12 | 77.55 |
| | | D4 | 56.3 | 18 | 84.69 |
| Mean Average Precision (MAP) | $\frac{TP+TN}{TP+FN+TN+FP}$ | Dataset | Score | AVG FPS | Accuracy |
| | | D1 | 48.9 | 17 | 81.25 |
| | | D2 | 41.2 | 12 | 77.65 |
| | | D3 | 39.3 | 13 | 73.75 |
| | | D4 | 47.6 | 16 | 81.01 |
| F1 | $2.\frac{Precision.Recall}{Precision+Recall}$ | Dataset | Score | AVG FPS | Accuracy |
| | | D1 | 57.20 | 22 | 87.80 |
| | | D2 | 45.36 | 15 | 85.31 |
| | | D3 | 43.84 | 14 | 83.23 |
| | | D4 | 55.59 | 12 | 85.21 |

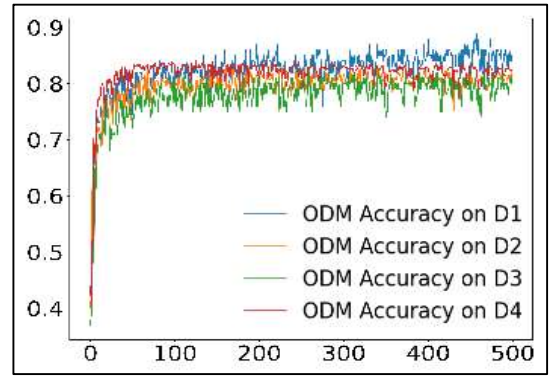

Fig. 13. Accuracy of ODM

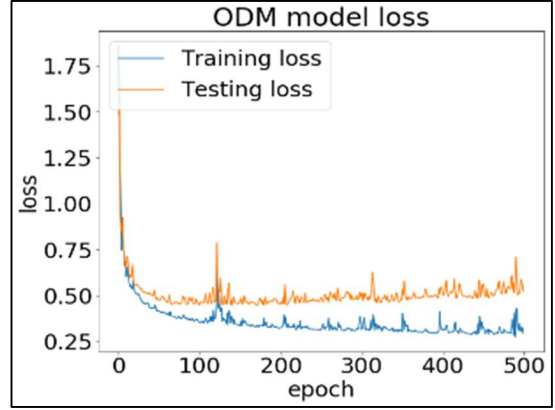

Fig. 14. Loss of ODM

IV. DISCUSSION

Starting with the development of 3-Survivor, it is expected that the technologies of this proposed robot will develop. Analyzing the operational results, it can be evaluated that 3-Survivor can be further upgraded to adopt different functionalities that serve both as a rescue and service robot. One of the major drawbacks observed from the operation is the shape and outflow of the actuator mechanism. In particular, for a heavy rescue operation, the actuator plays an important role in object rescuing. Thus, the actuator needs to be upgraded. Although 3-Survivor is designed to operate small rescue operations, the accuracy of the ODM and body design outperformed the previous versions and other state-of-art robots. On the other hand, additional features like autonomous movement, strong communication system, humanoid body shape can improve the functionality of the robot. In the future version of 3-Survivor, the mentioned lacking's will be handled.

## V. CONCLUSION

In this work, the design, analysis, and development of 3-Survivor are presented. A passive double-tracked mechanism has been constructed for rescue missions in a hazardous environment. The body configuration with parameters is acquired by the mean of dynamic analysis of simulation and operational mechanism. A series of experiments displayed the overall accuracy and performance of field operations. It is evident that the operational functionalities of 3-Survivor show significant improvement from current studies. The ODM and body design is the most studied concern for this robot and the result from the experiments are reliable. In the future, a military and human assistant version of 3-Survivor will be introduced with various autonomous technologies to guarantee the reliability of the developed rescue robots.